\documentclass[runningheads]{llncs}
\usepackage{amsmath,amssymb} % define this before the line numbering.
\usepackage{color}
\usepackage[pdftex]{graphicx}
\usepackage{sidecap}
\usepackage{epsfig}
\usepackage{subfig}
\usepackage{hyperref}
\usepackage{appendix}
\usepackage{caption}
\usepackage[width=122mm,left=12mm,paperwidth=146mm,height=193mm,top=12mm,paperheight=217mm]{geometry}
\begin{document}
% \linenumbers
\pagestyle{headings}
\mainmatter
\def\ECCV16SubNumber{286}  % Insert your submission number here

\title{HDRFusion: HDR SLAM using a low-cost auto-exposure RGB-D sensor} % Replace with your title

%\titlerunning{ECCV-16 submission ID \ECCV16SubNumber}

\authorrunning{S. Li$^1$, A. Handa$^2$, Y. Zhang$^1$, A. Calway$^1$}

%\author{Anonymous ECCV submission}
\author{Shuda Li$^1$, Ankur Handa$^2$, Yang Zhang$^1$,\\
Andrew Calway$^1$}
\institute{$^1$University of Bristol, $^2$University of Cambridge}

\maketitle
\begin{abstract}
We describe a new method for comparing frame appearance in a frame-to-model 3-D mapping and tracking system using an low dynamic range (LDR) RGB-D camera which is robust to brightness changes caused by auto exposure. It is based on a normalised radiance measure which is invariant to exposure changes and not only robustifies the tracking under changing lighting conditions, but also enables the following exposure compensation perform accurately to
allow online building of high dynamic range (HDR) maps. The latter facilitates the frame-to-model tracking to minimise drift as well as better capturing light variation within the scene. Results from experiments with synthetic and real data demonstrate that the method provides both improved tracking and maps with far greater dynamic range of luminosity.

\end{abstract}

\section{Introduction}

Most existing methods for dense visual/RGB-D 3-D mapping and tracking
rely on the brightness constancy assumption, i.e. the brightness of
3-D points observed from different viewing positions is constant.
These can be categorized into using either a global or a local
constancy assumption. The former assumes that any two over lapping
frames from a sequence fulfil the condition~\cite{Newcombe_ICCV11},
whilst the latter requires only that consecutive frames
do\cite{Kerl_ICRA13,Whelan_IJRR15}. The global assumption enables
frame-to-model tracking which is known to accumulate less
drift~\cite{Newcombe_ISMAR11}, whilst the local assumption is easier to
meet in practice but means that the tracking is done frame-to-frame,
with a consequent increase in drift.

However both of the above assumptions are broken in reality when using
cameras equipped with automatic exposure (AE). AE is designed to map
the high dynamic range of scene luminance into a narrow range for display 
devices while remain suitable for the human eye. When the camera moves from
a bright to dark area, the exposure time is increased automatically so
that more light can be captured by the camera sensor and vice versa when the
camera moves from dark to bright regions. This breaks the global
assumption since images viewing the same scene area from different
viewing positions are seldom captured at the same auto-exposure. The
local assumption is more likely to be met as exposure usually changes
smoothly, but this assumption also breaks when video flickering
occurs. Video flickering artefacts, also known as brightness
fluctuation, happen when a camera moves across the boundary between a
bright and dark area or moves quickly back and forth between them: in
these scenarios, the exposure changes dramatically in a short period
of time and results in flickering. Turning AE off can ensure the
brightness constancy, but it is often undesirable since it leaves
bright areas over exposed and dark areas under exposed, leading to the
loss of important visual detail.

\begin{figure}[t]
\centering
\includegraphics[width=1.\textwidth]{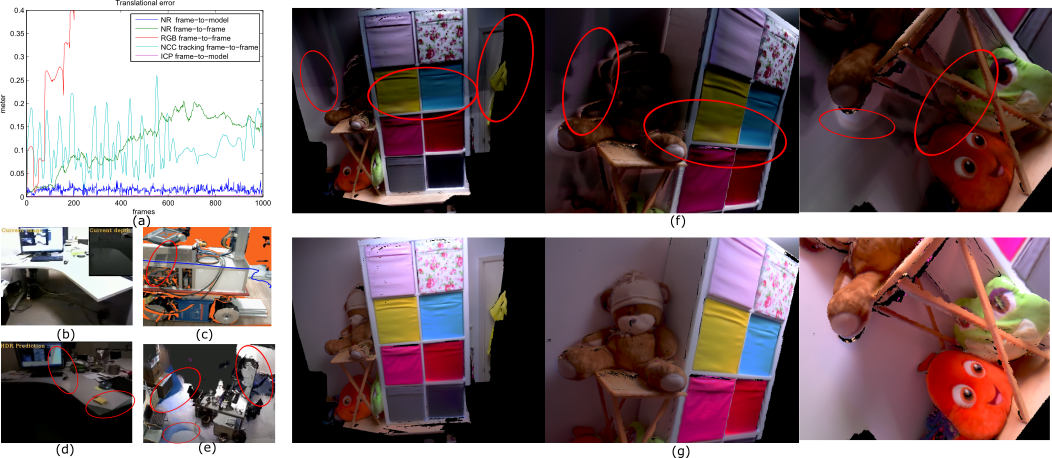}

\caption{(a) shows the proposed frame-to-model tracking using normalized radiance deliver
best tracking accuracy using visual data. The tracking is performed using a challenging synthetic flickering
RGB-D sequence. (b)-(e) are screen captures from video released with previous
works. Specifically, (b) and (d) are from \cite{Maxime_ISMAR13}, where (b)
is the raw input image. (d) is predicted scene textures.
By contrast, the unrealistic artefacts, marked by red circles, indicate
insufficent exposure compensation. (c) is predicted scene texture
from~\cite{Kerl2015}; (e) from~\cite{Whelan_RSS2015}. Similar
artefacts can be seen in these results. (f) in the top
right shows the results from our implementation of
\cite{Whelan_IJRR15} using a RGB-D video sequence, the artefacts are
very strong due to large camera exposure adjustment when moving from bright 
area (top in the scene) to the dark area (bottom left in the scene).
(g) in the bottom right are the predicted textures using
the proposed HDRFusion. It can been that it is free of artefacts and
its HDR textures are visualized using Mantiuk tone mapping operater~\cite{Mantiuk_TOG08}.}

\label{fig:res}

\end{figure}

AE also poses a problem when texturing a 3-D model of the scene.
Overlapping images captured with inconsistent brightness will leave
mosaic artefacts when projected back onto the model surface. This is a
common problem for many state-of-the-art dense mapping systems as
illustrated in Fig.~\ref{fig:res}. The problem has been widely
addressed in conventional model texturing, panoramic
imaging~\cite{Brown_IJCV07} and video tonal
stabilization~\cite{Aydin2014,Farbman2011}. These works tackle the
problem by compensating the global brightness of input images and
blending colours along the boundaries between input images to create
consistent texture. But these are usually expensive offline approaches
and are mainly aimed at delivering visually pleasing results rather
than maintaining fidelity to the real world luminance.

In this paper, we introduce a novel technique for appearance based
frame comparison which allows robust frame-to-model mapping and
tracking using RGB frames with AE enabled. It is very robust to
brightness fluctuation and is capable of capturing a consistent HDR
texture on the 3-D surface of the model (Fig.~\ref{fig:res}(a)). The HDR range corresponding
to real world luminance values is illustrated in
Fig.~\ref{fig:res}(j) using Mantiuk tone mapping operation
(TMO)~\cite{Mantiuk_TOG08}.

The key assumption of the work is that the luminance of real world is
globally constant and invariant to video brightness changes due to AE.
The main challenge lies in how to build a real-time system, capable of
tracking reliably with AE enabled so that HDR luminance can be
captured by fusing low dynamic range (LDR) frames together. Instead of jointly tracking and
compensating exposure like previous work~\cite{Maxime_ISMAR13} ---
which is not as robust and reliable as shown in our tests --- we
propose to track normalized radiance since it is a function which
depends on only luminance. Radiance is the amount of luminance
captured during the period of the exposure time. Another advantage of
tracking normalized radiance lies in that the normalization operation
can be efficiently implemented using down-sampled integral images.

Exposure compensation is therefore decoupled from tracking and greatly
improves its accuracy as well. In the end, both the tracking and
radiance fusion benefit from confidence maps derived from sensor noise
level function which adaptively weighs radiance map at pixel level.
Overall, the proposed HDRFusion achieves high quality radiance map and
enables better visualization experience using TMOs. We will
demonstrate the improvements in both qualitative and quantitative
experiments.

\begin{figure}[t]
\centering
\includegraphics[height=4.cm]{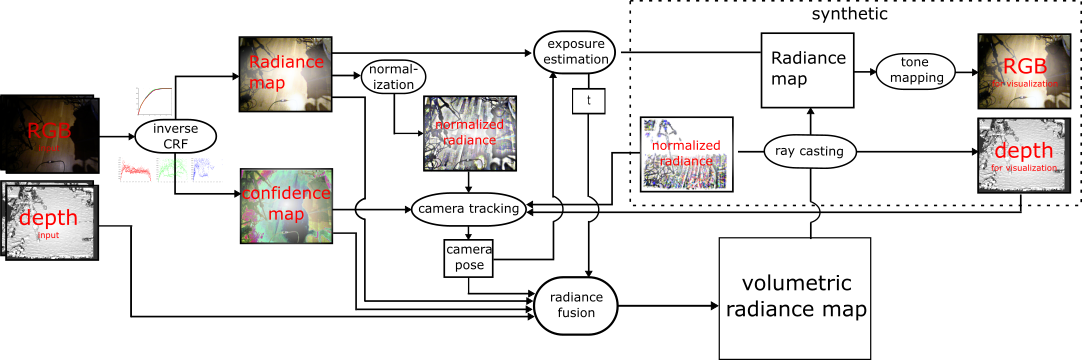}
\caption{Flow chart of HDRFusion. The boxes represent data structures, eclipses represent data transforming modules and arrows represent data flow. From left, input RGB frames are converted into radiance map. The camera pose is tracked in a frame-to-model style. Note that in confidence map is used in exposure estimation module but for simplitcity the data flow is not shown.}
\label{fig:overview}
\end{figure}

\subsection{Overview}
We now give an overview of HDRFusion. The main algorithm is shown in the flow-chart in Fig.~\ref{fig:overview}. The inputs are RGB-D frames from a Xtion sensor. Firstly, we estimate the inverse camera response function and noise level function for radiometric calibration (Section~\ref{sec:crf}). The RGB frames are then converted into radiance map with estimated pixel-wise confidence. The camera poses are tracked by aligning incoming frame with the prediction coming from the 3D model built so far, \textit{i.e.}, registering the live normalized radiance with predicted normalized radiance. The predicted normalized radiance is estimated by casting rays into the global volume. The confidence map is used to adaptively weigh error function for tracking, exposure compensation and radiance fusion. The ray casting module establishes predicted radiance, normalized radiance and depth. The predicted radiance and depth can be used for visualization through tone mapping on LDR devices or output as HDR data.

\section{Related work}
 
There is a huge wealth of literature on dealing with visual odometry or camera motion tracking. However, we will only focus on the direct approaches which can track and reconstruct a dense and textured 3-D model in real-time. Motion tracking using active sensor \cite{Newcombe_ISMAR11,Serafin_IROS15} is independent to lighting but leaves surfaces un-textured. Approaches combining appearance and depth \cite{Kerl_ICRA13,Whelan_ICRA13,Whelan_IJRR15,Kerl2015} for camera tracking are the most relevant approaches. In all these approaches, it is assumed that brightness of consecutive frames is constant which is likely to fail when video flickering happens. In addition, \cite{Whelan_IJRR15} introduce a simple color blending method but as shown in our experiments, it is inadequate to deal with large exposure changes. Kerl \textit{et al.} ~\cite{Kerl2015} introduce a key frame based approach by taking the rolling shutter effect into account. The approach relies on local brightness constancy when tracking live frames with a key frame, it is capable of producing sharp super-resolution frames involves no exposure compensation. 

\begin{figure}[t]
\centering
\includegraphics[width=1.\textwidth]{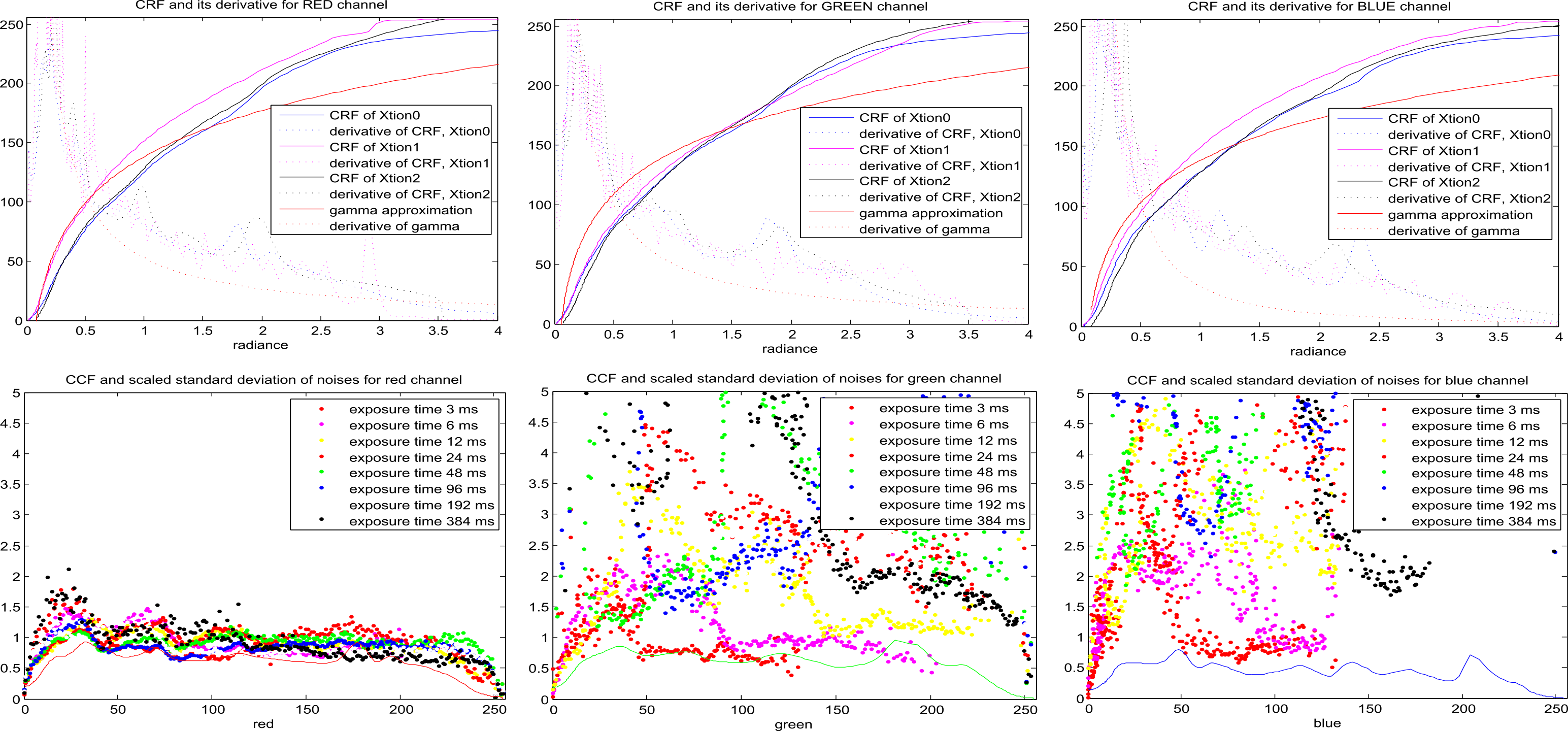}
\caption{The CRF and PCF of the RGB camera on 3 Xtion sensors. The figures in the top row are the CRF function and its derivative of RGB channels respectively. From the figure, we can see that the gamma approximation of CRF bears large error when the radiance is high. The figures in the bottom row are the PCF and scaled standard deviation of noise level captured as various exposure time. }
\label{fig:crf}
%\vspace{-1.5cm}
\end{figure}

Maxime \textit{et al.} \cite{Maxime_ISMAR13} propose one of the first work in real-time 3-D HDR texture capturing. We follow the same approach of transforming raw RGB into radiance domain and tracking using radiance. It mainly focuses on re-lighting virtual specular objects. The differences between~\cite{Maxime_ISMAR13} and this paper are two-fold. First, in \cite{Maxime_ISMAR13}, a gamma function is adopted to approximate inverse camera response function (CRF). Gamma function may introduce error when radiance is high and the resulting radiance is not directly proportional to scene luminance (Fig.~\ref{fig:crf}).  Second, in \cite{Maxime_ISMAR13}, the exposure is estimated jointly with camera pose, but we find that the shape of error function when tracking using exposure compensated radiance bears shallow global minima even when exposure has been compensated for and, therefore, not as robust as normalized radiance based object function we proposed.(Fig.~\ref{fig:track}) Lastly, mosaic artefacts are clearly visible from the synthetic HDR mode which indicate inadequate exposure estimation (Fig.~\ref{fig:res}(d)). 

Normalized cross correlation (NCC) has been widely applied in visual tracking~\cite{Scandaroli2012} to deal with challenging lighting condition but its computational cost grows exponentially with the size of patch. Small patches are sensitive to image noise and bring many local minimum (Fig.~\ref{fig:track}). In addition, the 3-D HDR texture capturing is not addressed in the paper. HDR video capture using high-end stereo rig~\cite{Heo_PAMI11,Batz2014} is also relevant to the topic since it involves estimating disparity between binocular views so that LDR frames captured by both frame can be integrated into a single stream of HDR video~\cite{Batz2014}, but the high quality HDR video is the main focus of the group of approach rather than a full 3-D model.  

\section{Preliminaries}

Start from direct tracking using visual data assuming brightness constancy, camera poses can be estimated by minimizing the intensity difference between a reference frame and a live frame. The object funciton $F$ can be formulated as:
\begin{equation}
F(\mathbf{R},\mathbf{t}) = \int_{\Omega}\left \|I_r(\mathbf{u}) - I_l(\pi(\mathbf{R}\pi^{-1}(\mathbf{u},D_r(\mathbf{u}))+\mathbf{t}))  \right \|^2_2d\mathbf{u} \label{eqn:photo}
\end{equation}

where $I:\Omega\rightarrow \mathbb{R}_+$ and $D:\Omega\rightarrow \mathbb{R}_+$ denote the intensity and depth functions. The whole 2-D image domain is denoted as $\Omega\subset \mathbb{R}^2$ and $\mathbf{u}\in \mathbb{R}^2$ is the pixel coordinate. Subscript $_r$ and $_l$ denote the reference frame and live frame respectively. $\mathbf{R}\in\mathbb{SO}(3)$ and $\mathbf{t}\in\mathbb{R}^3$ are the rigid body motion to transform a 3-D point defined in reference coordinate system to live coordinate system. $\pi:\mathbb{R}^3 \rightarrow \Omega$ and $\pi^{-1}:\Omega\times\mathbb{R}_+ \rightarrow \mathbb{R}^3$ are projection function and its inverse. 
$\pi(.)$ projects a 3-D point to image plane and $\pi^{-1}(.)$ transforms 2-D point back into 3-D given the depth $D$. 

Equation~\ref{eqn:photo} works as long as brightness constancy holds. We can define the correspondent point in live frame as $\mathbf{u}'=\pi(\mathbf{R}\pi^{-1}(\mathbf{u},D_r(\mathbf{u}))+\mathbf{t})$ and $e(\mathbf{u},\mathbf{u}') = I_r(\mathbf{u})-I_l(\mathbf{u}')$. Equation~\ref{eqn:photo} is rewritten as $F(\mathbf{R}, \mathbf{t})=\int_{\Omega} \left \|e(\mathbf{u},\mathbf{u}') \right \|^2d\mathbf{u}$. NCC based tracking can be viewed as an extension from equation~\ref{eqn:photo} by replacing $e(\mathbf{u},\mathbf{u}')$  with $\sqrt{1 - C^2(\mathbf{u},\mathbf{u}')}$. $C(.)$ is the NCC score and defined as following:

\begin{align}
C(\mathbf{u},\mathbf{u}')=& \frac{1}{|\Omega_N|^2}\int_{\Omega_{N}}\frac{(N_r(\mathbf{u},\mathbf{v})-\mu)(N_l(\mathbf{u}',\mathbf{v})-\mu')}{\sigma \sigma'} d\mathbf{v} \label{eqn:ncc}
\end{align}
Where $N:\Omega\times\Omega_N\rightarrow \mathbb{R}_+$ defines a small image patch, a neighbourhood centred at $\mathbf{u}$. $\Omega_N\subset\mathbb{R}^2$ is the domain of the neighbourhood $N$ and $\mathbf{v}\in\mathbb{R}^2$ is the coordinate w.r.t. $N$. $\mu$ and $\sigma$ are mean and std. (standard deviation) of image intensity over $N_r$ and $\mu'$ and $\sigma'$ are mean and std. over $N_l$. The NCC-based tracking can be formulated as $F(\mathbf{R}, \mathbf{t})=\int_{\Omega} \left \|1-C^2(\mathbf{u},\mathbf{u}') \right \|d\mathbf{u}$.

\section{Camera imaging process}
\label{sec:crf}
The key observation we rely on in this paper is that the scene luminance is mostly constant and invariant to exposure settings. The idea is to replace $e(.)$ with a new error function dependent on luminance only. The luminance $L$ is the radiance $R$ received at the camera sensor per unit time $L=R/\Delta t$, where $\Delta t$ is the exposure time. The relation between luminance and image intensity $I$ can be described by the image formation model~\cite{Hasinoff2010}:

\begin{align}
   I &= f(R + n_s(R) + n_c) 
\end{align}
where $f:\mathbb{R}_+\rightarrow \mathbb{Z}_+$ is the camera response function (CRF) and $R=L\Delta t$. Essentially, it maps radiance $R$ to LDR intensity level $I$, which is ranged from 0 to 255.  $n_s$ accounts for noise component dependent on the radiance, $n_c$ accounts for the constant noise. The statistics of noise can be assumed as $E(n_s) = E(n_c) = 0$, $Var(n_s)=L\Delta t \sigma_s^2$ and $Var(n_c)=\sigma_c^2$. 

The noise level function~\cite{Liu_PAMI08,Best_ECCV12} measures how reliable sensor response is at given intensity level. For convenience, we convert it to a probability function by scaling the noise level function using a scalar $m$, where $m$ is the maximum standard deviation over 3 colour channels. 
\begin{align}
   p(I) &= \frac{1}{m}\left.\frac{\partial{f(r)}}{\partial{r}}\right\vert_{r=R} \sqrt{R\sigma_s^2+\sigma_c^2}\label{eqn:ccf}
\end{align}
where $p:\mathbb{Z}_+\rightarrow (0,1)$. $R=f^{-1}(I)$ is the radiance. In the right column of Fig.~\ref{fig:nr}, the probability maps are shown. Each channel represents the probability of the channel at the pixel location: dark areas show the low probability pixels which usually occur around exposed or under exposed parts of the image. We can also define variants of this probability function based on equation~\ref{eqn:ccf}. $p_0(I)=1$, $p_i(I)=\sqrt{p(I)}$, $p_2(I)=p(I)$, and $p_3(I)=p(I)^2$. For clarity, the family of probability functions are named as pixel confidence functions (PCF) from now on since these are different from noise level functions. Their effects will be discussed in section~\ref{sec:nr}.

The CRF and PCF depends on specific type of camera sensor. They can be pre-calibrated before performing the HDRFusion~\cite{Debevec1997,Kim_CVPR04,Kim_CVPR07}. Specifically, our CRF is estimated by putting the RGB-D sensor at fixed position. A sequence of images at different exposure time are captured ~\cite{Debevec1997} and the noise level function and PCF are estimated using~\cite{Best_ECCV12}. The CRF, its derivative and PCF are shown in Fig.~\ref{fig:crf}. With this estimated CRF, its inverse $f^{-1}(.)$ and PCF can be calculated straightforwardly: inverse CRF, allows us to convert intensity to radiance efficiently and PCFs allow us to weigh the error terms appropriately in tracking, exposure compensation and fusion stage. 

\begin{figure}[t]
\centering
\includegraphics[width=1.\textwidth]{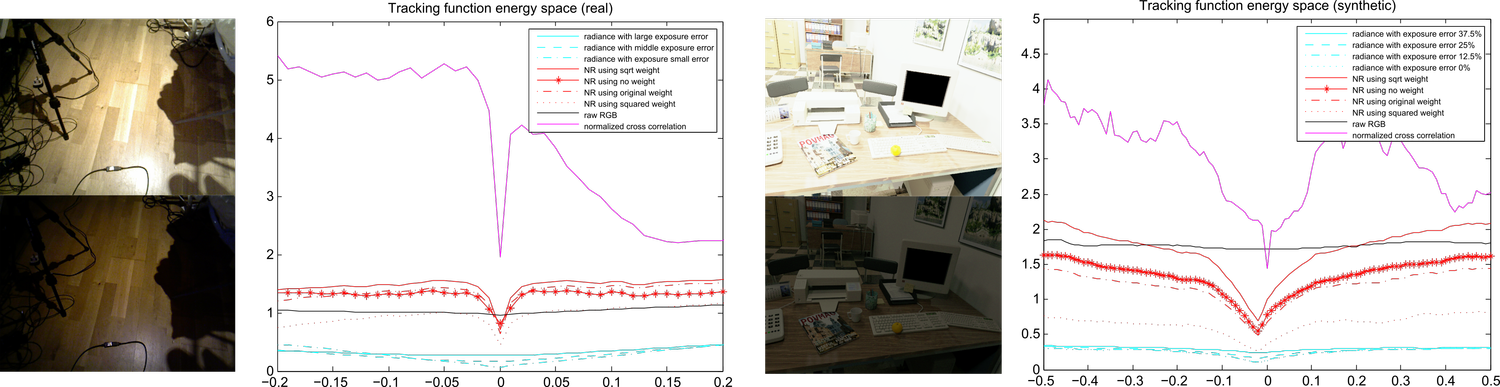}
\caption{The comparison of tracking errors. The family of error function using normalized radiance (red) gives the most ideal global minimum over the ground truth. The NCC based error function~\cite{Scandaroli2012} also presents a strong convex but bears a lot of local minimum. Tracking using exposure compensated radiance~\cite{Maxime_ISMAR13} looks better than tracking raw RGB but its global minimum are shallow even when the exposure is compensated with high accuracy. The left plotting uses real flickering pair and the right uses simulated flicking pairs based on~\cite{Best_ECCV12}. }
\label{fig:track}
\end{figure}

\section{Normalized radiance}\label{sec:nr}

\begin{figure}[t]
\centering
 \subfloat{%
      \includegraphics[width=0.15\textwidth]{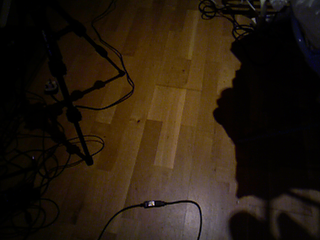}
    }
		\subfloat{%
      \includegraphics[width=0.15\textwidth]{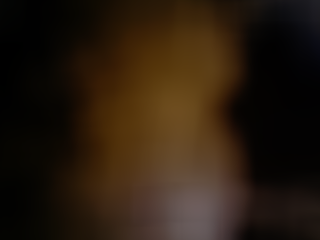}
    }
		\subfloat{%
      \includegraphics[width=0.15\textwidth]{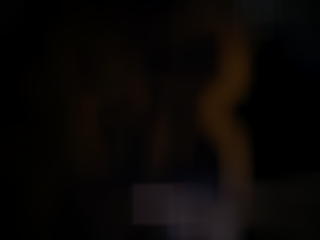}
    }
		\subfloat{%
      \includegraphics[width=0.15\textwidth]{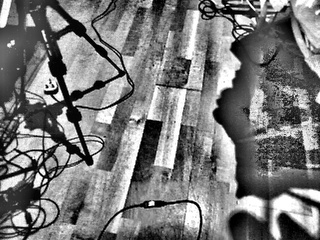}
    }
		\subfloat{%
      \includegraphics[width=0.15\textwidth]{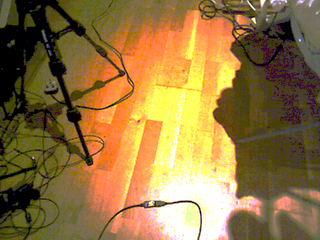}
    }	\\
	\vspace{-.25cm}
 \subfloat{%
      \includegraphics[width=0.15\textwidth]{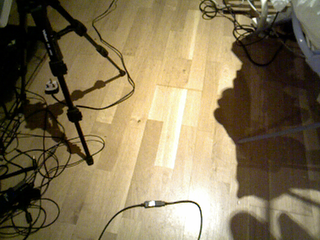}
    }
 \subfloat{%
      \includegraphics[width=0.15\textwidth]{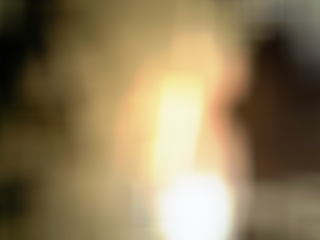}
    }
  \subfloat{%
      \includegraphics[width=0.15\textwidth]{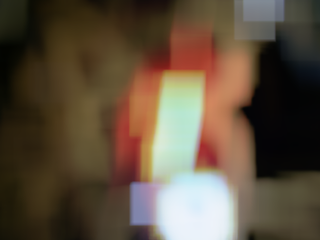}
    }
		\subfloat{%
      \includegraphics[width=0.15\textwidth]{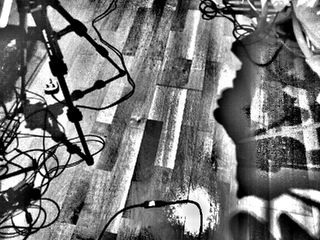}
    }
		\subfloat{%
      \includegraphics[width=0.15\textwidth]{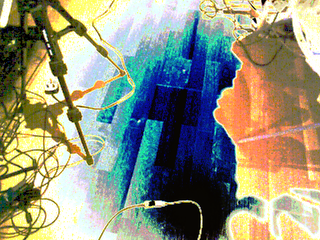}
    }		
 \caption{Radiance normalization. From left to right, figures correspond to raw RGB, mean, standard deviation, normalized radiance, and confidence map. The $1^{st}$ and $2^{nd}$ row correspond to 2 consecutive frames when flickering happens. Although the image brightness changes significantly, the normalized radiance map is pretty similar thank to equation~\ref{eqn:nr}. The mean, standard deviation maps and normalized radiance are tone mapped from HDR.}
 \label{fig:nr}
\end{figure}

Now we derive a novel error function dependent on luminance alone. The normalization of the radiance in a patch of neighbourhood $N$, centred at pixel location $\mathbf{u}$, can be formulated in the following:
\begin{align}
   \bar R_N(\mathbf{u}) &= \frac{R_N(\mathbf{u})-E(R_N)}{\sqrt{Var(R_N)}} = \frac{L_N(\mathbf{u})\Delta t - E(L_N\Delta t)}{\sqrt{Var(L_N\Delta t)}} = \frac{L(\mathbf{u})-E(L_N)}{\sqrt{Var(L_N)}} \label{eqn:nr}
\end{align}
where $R_N:\Omega_N \rightarrow \mathbb{R}_+$ and $L_N:\Omega_N \rightarrow \mathbb{R}_+$ denote radiance map and luminance map in $N$, respectively. From above equation, it can be seen that $\bar R_N(\mathbf{u})$ is independent from exposure $\Delta t$. This value is also invariant to viewpoint due to the fact that the luminance distribution in the local region corresponding to $N$ is roughly constant to viewing position, as long as the surface is Lambertian. Fig.~\ref{fig:nr} shows the mean, standard deviation, normalized radiance and confidence map of two consecutive frames captured at different exposure time. It can be seen that the normalized radiance maps extracted from frames captured at different exposure are strikingly similar while the mean and standard deviation maps are smooth and blurry which indicates good resistance to viewpoint changes. Therefore, the new error function can be defined as: 
\begin{align}
e'(\mathbf{u},\mathbf{u}')=(\bar R_r(\mathbf{u}) - \bar R_l(\mathbf{u}'))p(I_l(\mathbf{u}')) \label{eqn:ne}
\end{align}
where the probability $p(I_l(\mathbf{u}'))$ serves as a dynamic weight to balance the noise introduced during image formation such that less reliable pixel will be assigned with a smaller weight. $p(.)$ can be chosen from the family of PCFs we defined before. $p(.)\in\{p_0(.), p_1(.), p_2(.), p_3(.)\}$. 

The error functions using NCC, raw intensity, radiance with exposure compensated and the proposed normalized radiance are compared by plotting against the ground truth along x-axis in Fig.~\ref{fig:track}. Pairs of flickering consecutive frames are chosen, where one is real and the other is synthetic.  It can be seen that our proposed error function using normalized radiance and weighted by square root PCF $p_1(.)$ gives the most ideal error space for optimization. 

The camera poses can then be solved out by optimizing the error functions using the forward compositional approach described in~\cite{Whelan_IJRR15}. 

\subsection{Exposure compensation}
When the camera pose is estimated, the exposure will then be compensated using the follow equation:  
\begin{align}
t = \frac{1}{|\Omega|}\int_{\Omega} p_l(\mathbf{u})\frac{R_r(\mathbf{u})}{R_l(\mathbf{u'})} d\mathbf{u}
\end{align}
where  $p_l(.)$ is the PCF of live frame. After $t$ is estimated, the radiance map of live frame will be scaled by $t$. 

\section{Radiance Fusion}
\begin{figure}[ht]
\centering
\includegraphics[height=3.cm]{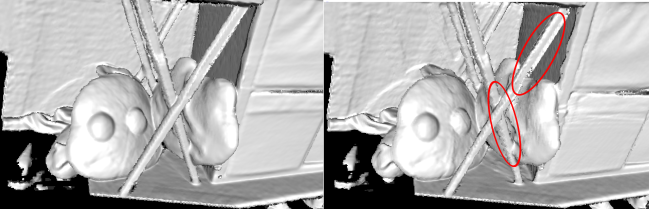}
\caption{Weight TSDFs according the angle between viewing direction and surface normal improves the geometry quality around thin and corner structures.}
\label{fig:fusion}
\end{figure}

The exposure compensated radiance map $tR_l$ will then be fused into a global volume using an fast parallel approach similar to~\cite{Whelan_IJRR15}. The volumetric data structure stores not only the truncated signed distance function (TSDF) and its weights, but also the 3 channels of radiance and normalized radiance and radiance weights. The normalized radiance is also fused into the global volume so that synthetic normalized radiance map can be efficiently extracted using ray casting. Note that the radiance weight is different from TSDF weights. The fusion of radiance with depth for each voxel is shown in the following equations:

\begin{align}
F &= \frac{w_F * F + w_F'*F'}{w_F + w_F'}\label{eqn:tsdf}\\
R &= \frac{w_R * R + w_R'*R'}{w_R + w_R'}\label{eqn:radiance}\\
\bar R &= \frac{w_R * \bar R + w_R'*\bar R'}{w_R + w_R'}\label{eqn:radiance}\\
w_F&= w_F + w_F'\\
w_R&= w_R + w_R'
\end{align}

where $F$ and $R$ are TSDF values and radiance in global volume; $F'$ and $R'$ are those from live frame. Similarly, $w_F$ and $w_R$ are the global weights. $w_F'$ and $w_R'$ are weights from live frame. $w_F = |\mathbf{n}^T\mathbf{v}|$ is the absolute cosine values between surface normal $\mathbf n$ and viewing direction $\mathbf v$ at the live pixel location where $\mathbf{n},\mathbf{v}$ are unit vectors. It down weight the TSDF values captured at high angle between the normal and viewing direction. Its effect is illustrated in Fig.~\ref{fig:fusion} $w_R= \frac{p_r+p_g+p_b}{3}$, where $p_r$, $p_g$ and $p_b$ are the PCF values of 3 colour channels respectively. In experiments, we find that storing individual PCF of 3 colour channel is the global volume is unnecessary and may introduce color distortion as well. 

To ensure the quality of radiance, only the pixels whose maximum PCFs are above a threshold $\tau_0$ and the angle between surface normal and viewing direction is above threshold $\tau_1$ are allowed to be fused into the volume. 
\begin{align}
\{R|max(p_0, p_1, p_2)>\tau_0)\bigcap \mathbf{n}^T\mathbf{v}<\tau_1\}
\end{align}

\section{Experiments}

In all experiments, we have used 3 Xtion RGB-D sensors whose exposure can be specified. Calibrated CRFs of them are plotted in Fig.~\ref{fig:crf}. Except CRF and noise level function, no other parameters need to be calibrated.  Camera intrinsics are set as default values as in OpenNI library. A C++ implementation and testing data for both the main HDRFusion and its calibration are available in \footnote{ \url{https://lishuda.wordpress.com}}. The codes are tested on two commodity system, PC0 equiped with NVIDIA GTX 680 and PC1 NVIDIA GTX Titan Black GPU. Both PC are hosted by an i7 quad-core CPU. The volume resolution are set as $256^3$ and $480^3$ for PC0 and PC1 respectively with volume size ranges from $2^3$ to $3^3$m according to the size of the scene. Frame resolution are set as QVGA for PC0 and VGA for PC1. Both of them operates at about 10Hz. We present a qualitative comparison with \cite{Whelan_IJRR15} and demonstrate the quality or recover HDR radiance map in an accompanying video: \url{https://youtu.be/ehwiFkmFQ7Q}.

\subsection{Tracking under flickering}
\begin{figure}[ht]
\centering
\includegraphics[width=1.\textwidth]{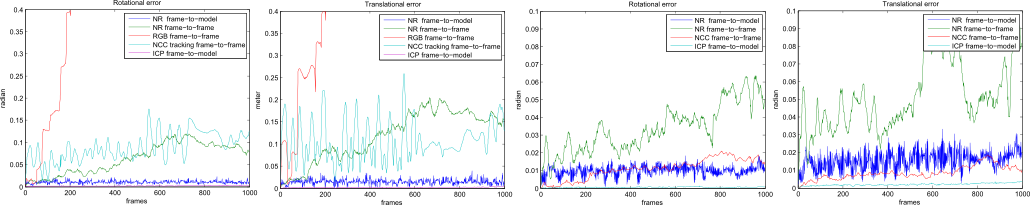}
\caption{Tracking synthetic sequence. The left two figures are the rotational and translational error using synthetic flickering sequence. The right two figures are using synthetic smooth AE sequence. In the flickering sequence, we can see that raw RGB based tracking quickly get lost, while the NCC and the proposed frame-to-frame tracking (NR) and frame-to-model tracking using normalized radiance remains working well. The tracking NR in frame-to-model mode gives the best performance in the flickering sequence. Due to the rich geometric variance, the ICP-based frame-to-model tracking give the best results. In smooth sequences, the ncc and ICP performs better but the proposed tracking remain working reasonably accurate. The frame-to-model tracking is within 3cm meter in the 1000 frames testing sequence.}
\label{fig:eval}
\end{figure}

We first use synthetic dataset ICL to evaluate our approach~\cite{Best_ECCV12}. The high quality CG HDR frames and ground truth camera poses are available. First, photo realistic LDR RGB frames are simulated using real CRF and noise level function of a randomly chosen Xtion sensor. We generates two sequences of video to simulate video flickering and smooth AE behaviour. The flickering sequence is simulated by randomly choosing exposure time from the set ${3,6,12,...,96}$ (ms). The second sequence is generated using the equation $\Delta t=C/L$, where $C=4.8\times10^5$ and $L$ is the average HDR intensity of the 10 by 10 patch in the center of the original HDR frames. The exposure simulated in the second way are changing smoothly. The Kinect like depth noise is also added using the approach from~\cite{Handa2014}. Typical flickering pairs are illustrated in Fig.~\ref{fig:track}. The tracking approach using normalized intensity, NCC object function based on~\cite{Scandaroli2012} and approach similar to the tracking of ~\cite{Maxime_ISMAR13} are used as baseline approaches. For fairness, the ICP-based frame-to-model tracking are disabled for all above methods. The tracking accuracy in terms of rotational and translational error are plotted in Fig.~\ref{fig:eval}.

\begin{figure}[ht]
\centering
\subfloat{%
      \includegraphics[width=0.25\textwidth]{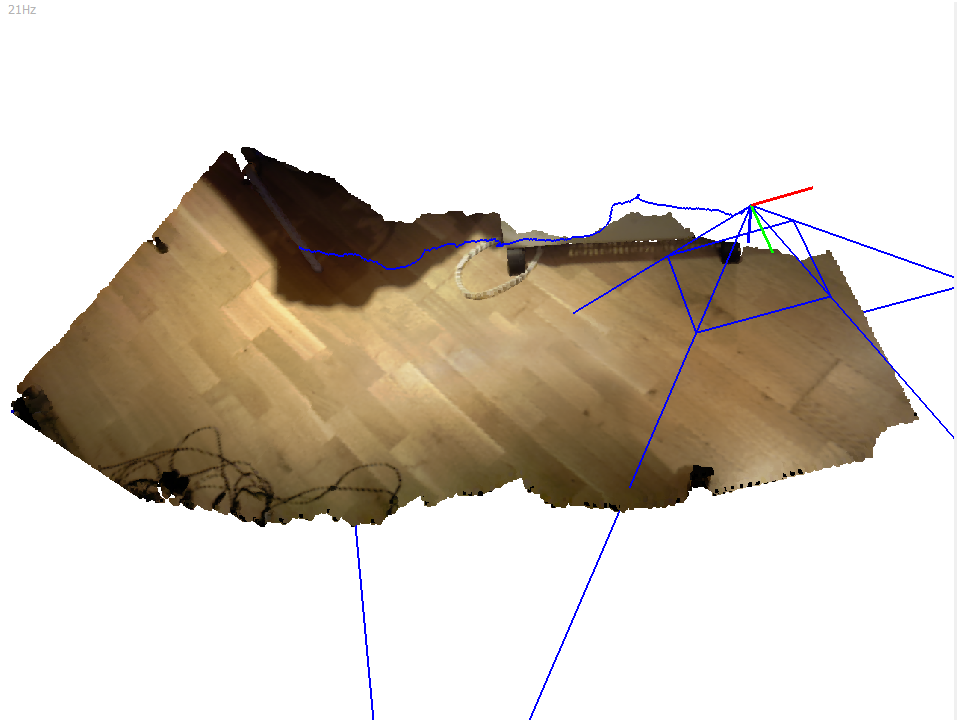}
    }
		\subfloat{%
      \includegraphics[width=0.25\textwidth]{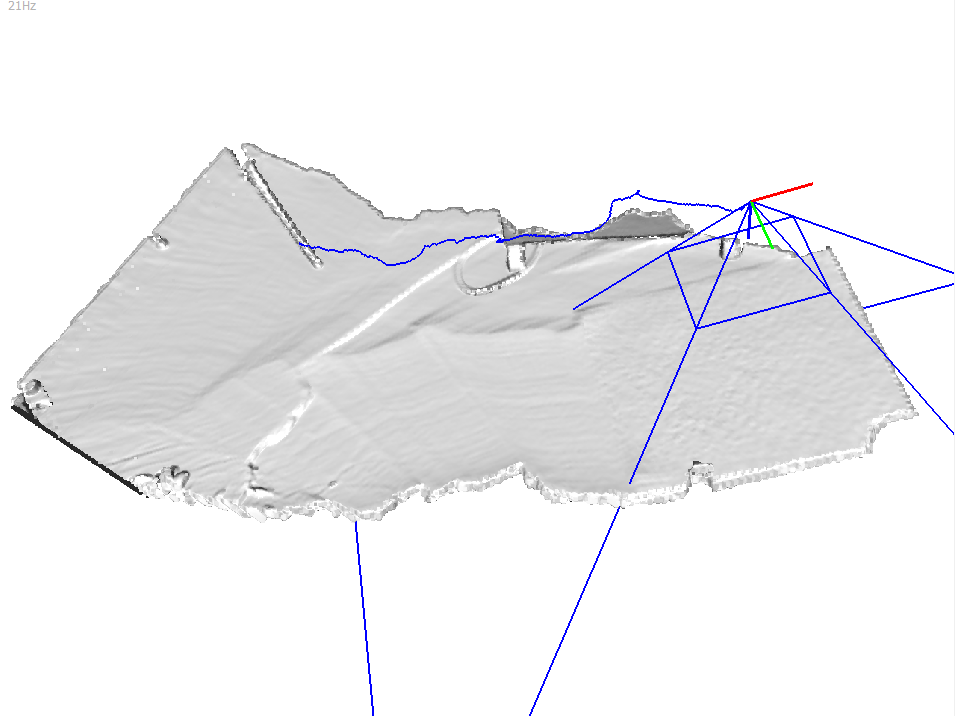}
    }
				\subfloat{%
      \includegraphics[width=0.25\textwidth]{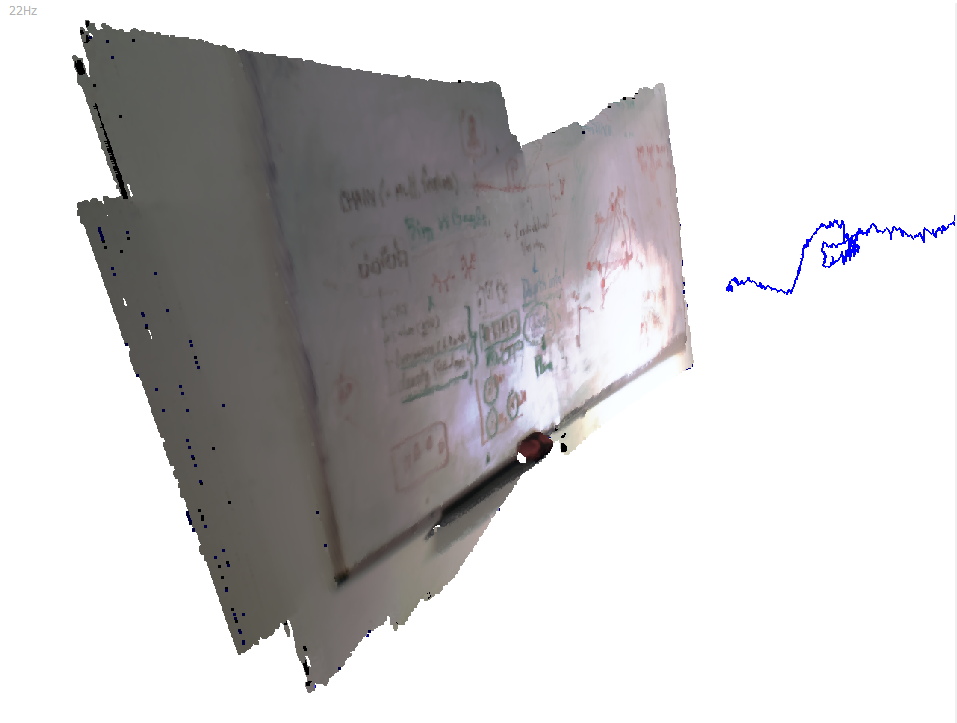}
    }
				\subfloat{%
      \includegraphics[width=0.25\textwidth]{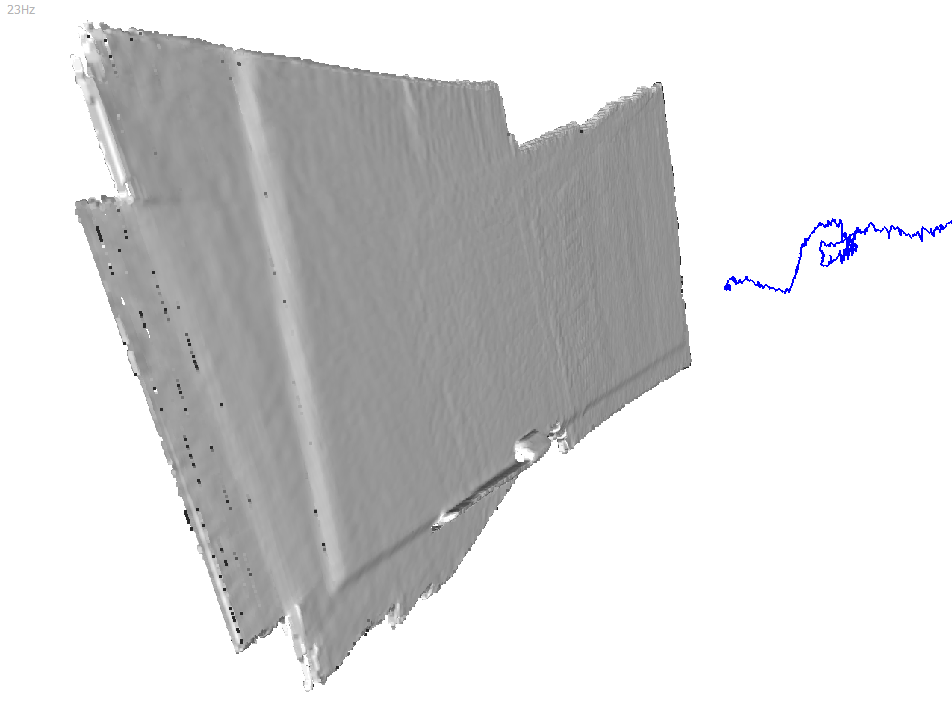}
    }
\caption{Tracking under flickering using real data. The blue curves are the camera trajectories. The frustums in the left figure show the camera pose.}
\label{fig:real}
\end{figure}

We also performed a qualitative comparison using real data between the proposed tracking and tracking using the approach from~\cite{Whelan_IJRR15}. Two sequences of RGB-D video with flickering are captured. In these sequences, the sensor is overlooking a floor and a white board respectively. As the camera moving from dark to bright areas, video flickering happens. ~\cite{Whelan_IJRR15} fails to tracking when flickering happens, while the proposed method remain tracking effectively. The reconstructed floor and white board using proposed approach are shown in Fig.~\ref{fig:real}. The tracking comparison between our approach and~\cite{Whelan_IJRR15} is also available in the accompanying video.

\subsection{HDR Radiance map}

\begin{figure}[ht]
\centering
\includegraphics[width=.7\textwidth]{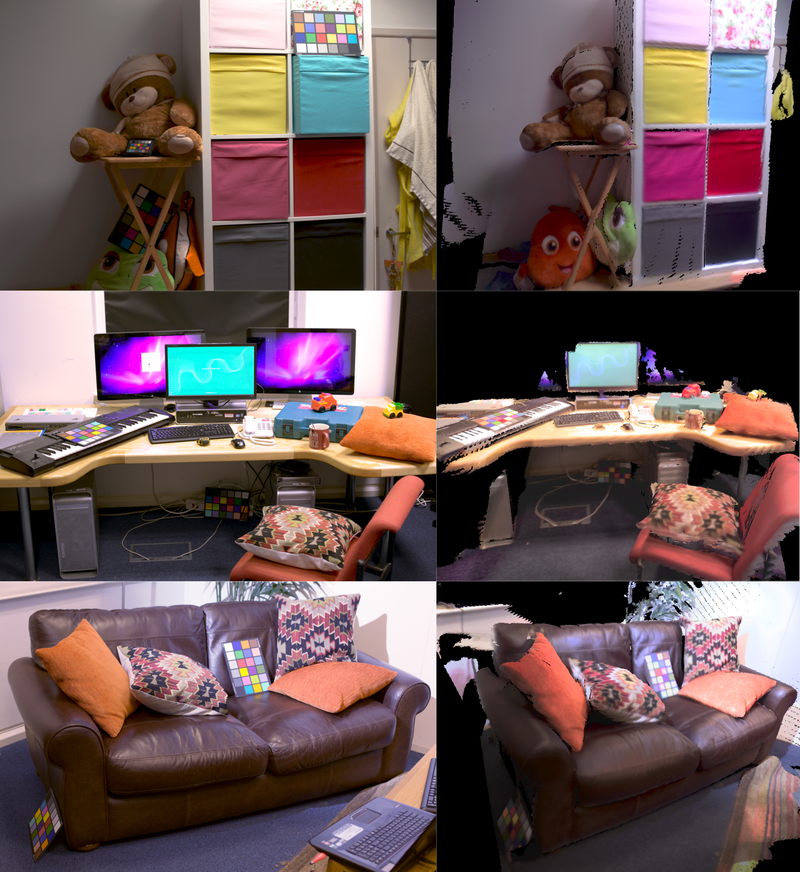}
\caption{The left are the ground truth HDR radiance and HDR radiance generated using HDRFusion are rendered using ~\cite{Mantiuk_TOG08} where the colour saturation is set as 1. We can see that estimated HDR texture closely matches the HDR radiance captured using the high-end SLR camera.}
\label{fig:canon}
\end{figure}

\begin{figure}[h]
\centering
\includegraphics[width=.8\textwidth]{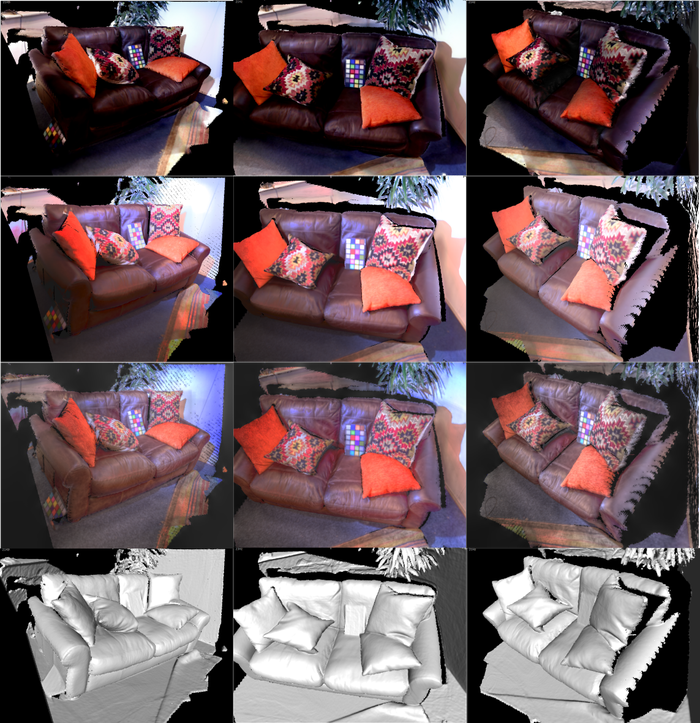}
\caption{Sofa. The LDR frames generated using ~\cite{Whelan_IJRR15} are shown in the first row and HDR frames produced by HDRFusion are shown in the second row and third row. The second row is generated using ~\cite{Mantiuk_TOG08} where the colour saturation is set as 1. Comparing with raw RGB fusion~\cite{Whelan_IJRR15}, the dynamic range of the radiance texture is much higher. The details in dark area are well preserved. The third row is generated using~\cite{Mantiuk2006}, where the colour saturation is set as 1.25. ~\cite{Mantiuk2006} visualizes the rich details captured by HDRFusion. The bottom row shows the recovered surface geometry.}
\label{fig:hdr3}
\end{figure}

\begin{figure}[t]
\centering
\includegraphics[width=.9\textwidth]{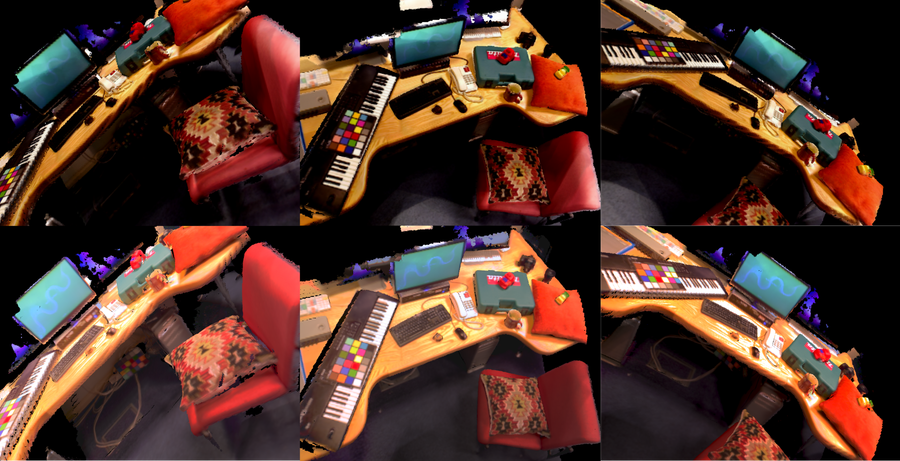}
\caption{Desk. The LDR frames generated using ~\cite{Whelan_IJRR15} are shown in the first row and HDR frames produced by HDRFusion are shown in the second row. The HDR radiance is rendered using~\cite{Mantiuk_TOG08}, where the colour saturation is set as 1.5. The luminance under the desk is very low but are well preserved in the HDR radiance map.}
\label{fig:hdr4}
\end{figure}

The HDR radiance are shown both in the screen shots attached in the paper and in the accompanying video. In Fig.~\ref{fig:res},~\ref{fig:canon}, ~\ref{fig:hdr3} and~\ref{fig:hdr4}, we perform the proposed HDRFusion in three scenes, namely 'Bear', 'Desk' and 'Sofa'. The bear sequence is illuminated by indirect sun light. The desk sequence is illuminated by Fluorescent. The sofa sequence is illuminated by both fluorescent lighting and Dedolight-400D metal halide lamp. In Fig.~\ref{fig:canon}, HDR scene textures are compared with the ground truth. The ground truth is captured using a Canon 5D MarkII SLR camera. Three exposure LDR images with a 2-fstop interval of the scene were captured and then merged to form an HDR image. Both are rendered using tone maping operator(TMO)~\cite{Mantiuk_TOG08}. 

\section{Conclusion}
In this paper, we propose a novel HDRFusion system capable of capturing high quality HDR scene texture using a low cost RGB-D sensor. Tracking normalized radiance allows decouple the tracking from exposure compensation which improves the accuracy of both. Tracking normalized radiance is also shown to be robust to video flickering due to camera AE adjustment. The tracking is runing in frame-to-model mode which accumulates less drift. In future work, calibrating the CRF function online will be investigated as in some sensors the exposure time can not be changed by user. Another limitation of the system lies in its large memory footprint. Storing both the normalized radiance and radiance seems unnecessary. Reducing the size of memory cost by combining the both will also be investigated.

%\vspace{-0.1cm}

%\vspace{-5.1cm}

\clearpage

\bibliographystyle{splncs}
\bibliography{abbrev,library}
\end{document}